# Integrated Image and Location Analysis for Wound Classification: A Deep Learning Approach

Yash Patel[1, #], Tirth Shah[1, #], Mrinal Kanti Dhar[1], Taiyu Zhang[1], Jeffrey Niezgoda[2], Sandeep Gopalakrishnan[3], and Zeyun Yu[1,4*]

[1]Department of Computer Science, University of Wisconsin-Milwaukee, Milwaukee, WI, USA.
[2]Advancing the Zenith of Healthcare (AZH) Wound and Vascular Center, Milwaukee, WI, USA.
[3]College of Nursing, University of Wisconsin Milwaukee, Milwaukee, WI, USA.
[4]Department of Biomedical Engineering, University of Wisconsin-Milwaukee, Milwaukee, WI, USA.

[*]Corresponding authors: Email: yuz@uwm.edu

[#] Authors have an equal contribution.

## Abstract

The global burden of acute and chronic wounds presents a compelling case for enhancing wound classification methods, a vital step in diagnosing and determining optimal treatments. Recognizing this need, we introduce an innovative multi-modal network based on a deep convolutional neural network for categorizing wounds into four categories: diabetic, pressure, surgical, and venous ulcers. Our multi-modal network uses wound images and their corresponding body locations for more precise classification. A unique aspect of our methodology is incorporating a body map system that facilitates accurate wound location tagging, improving upon traditional wound image classification techniques. A distinctive feature of our approach is the integration of models such as VGG16, ResNet152, and EfficientNet within a novel architecture. This architecture includes elements like spatial and channel-wise Squeeze-and-Excitation modules, Axial Attention, and an Adaptive Gated Multi-Layer Perceptron, providing a robust foundation for classification. Our multi-modal network was trained and evaluated on two distinct datasets comprising relevant images and corresponding location information. Notably, our proposed network outperformed traditional methods, reaching an accuracy range of 74.79% to 100% for Region of Interest (ROI) without location classifications, 73.98% to 100% for ROI with location classifications, and 78.10% to 100% for whole image classifications. This marks a significant enhancement over previously reported performance metrics in the literature. Our results indicate the potential of our multi-modal network as an effective decision-support tool for wound image classification, paving the way for its application in various clinical contexts.

## Introduction

Wound diagnosis and treatment are a pressing issue worldwide, with a considerable population suffering from wounds. As per a 2018 retrospective analysis, the costs for wound treatment have been estimated to be between $28.1 billion to $96.8 billion [3], reflecting the tremendous financial and medical burden. The most commonly observed wounds include diabetic foot ulcer (DFU), venous leg ulcer (VLU), pressure ulcer (PU), and surgical wound (SW), each associated with a significant portion of the population [22][23][24][25]. Given these circumstances, effective wound classification is crucial for timely and adequate treatment.

Until recently, wounds were predominantly classified manually by specialists, often leading to inconsistencies due to lack of specific guidelines. However, the advent of artificial intelligence (AI) has brought about significant changes in healthcare, including wound diagnosis [4]. AI, specifically deep learning (DL), has proven to be a game-changer in medical image analysis, enabling accurate, time-efficient, and cost-effective wound classifications. Data-driven techniques like DL, which require minimal human intervention, have been extensively utilized for identifying patterns and relationships in complex data [5][11].

DL encompasses several methods, like Convolutional Neural Networks (CNN), Deep Belief Networks (DBN), Deep Boltzmann Machines (DBM), Stacked Autoencoders, and many more. These techniques have found applications in various medical diagnostic fields, including wound image analysis [26][29]. Studies have underscored the effectiveness and efficiency of deep convolutional neural networks in wound diagnosis and analysis [14][15][16].

Notwithstanding the advancements, the accuracy of wound classification models remains constrained due to the partial information incorporated in the classifiers. The present research introduces an innovative approach to address this limitation by including wound location as a significant feature in the wound classification process. Wound location, a standard entry in electronic health record (EHR) documents, is instrumental in wound diagnosis and prognosis. A body map has been utilized to facilitate accurate and consistent wound location documentation [28], enhancing the classifier's performance by providing a more holistic set of data for classification. The classifier trained on both image and location features outperforms those reliant solely on image data.

A simplified workflow of this study is shown in **Figure 1**. The developed wound classifier takes both wound image and location as inputs and outputs the corresponding wound class.

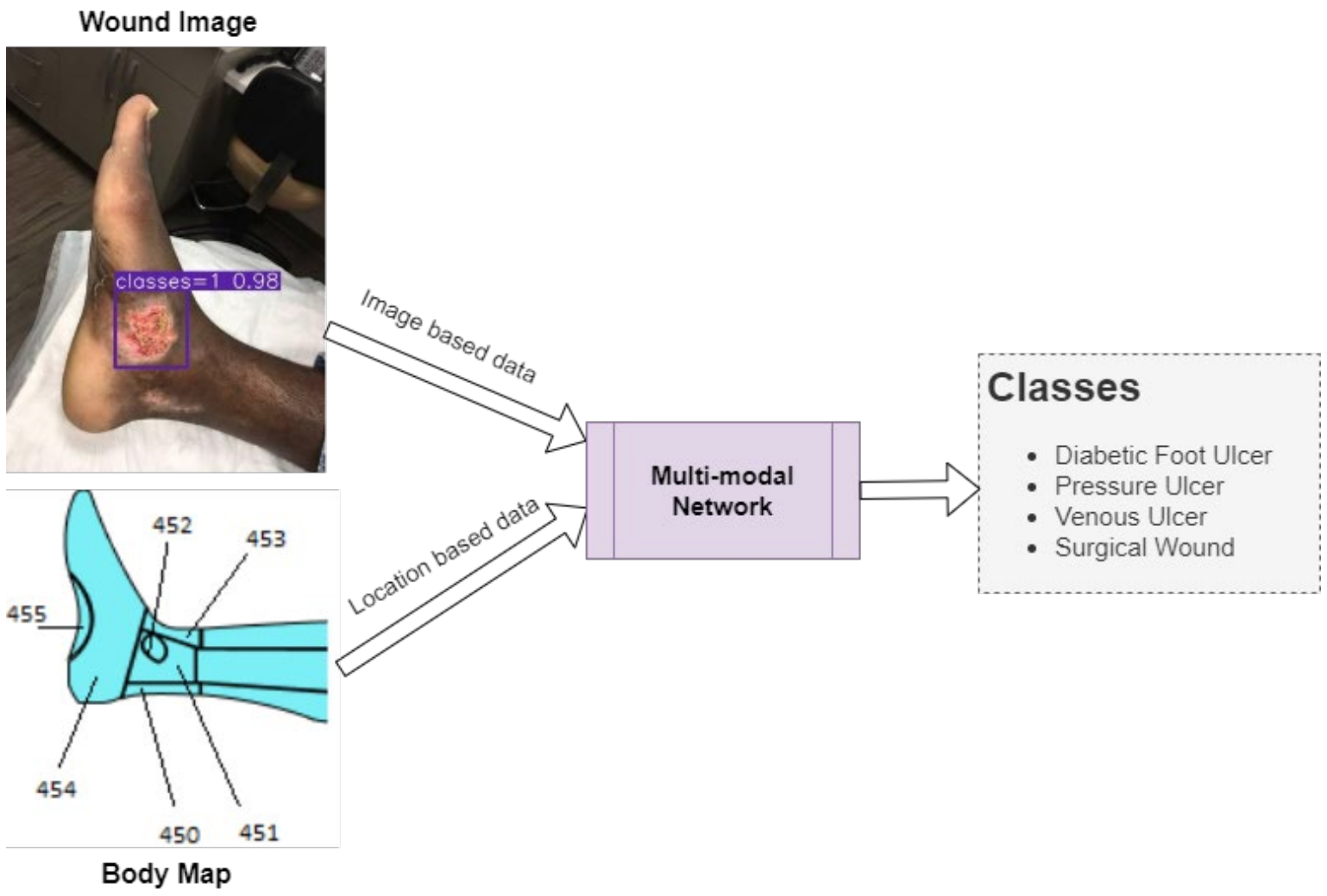

**Figure 1:** Expected workflow of this research

## Related Works

In this review, we revisit the relevant research in the field of wound image classification, segmented into categories based on the methodology of each study.

### A. Deep Learning Based Classification

**A.1. Convolutional Neural Networks (CNNs) with SVM:** A method proposed by Abubakar et al. [30] distinguished between burn wounds and pressure ulcers using pre-trained deep architectures such as VGG-face, ResNet101, and ResNet152 in combination with an SVM for classification. Similarly, Goyal et al. [31] predicted the presence of infection or ischemia in Diabetic Foot Ulcers (DFUs) using Faster RCNN and InceptionResNetV2 networks, in combination with SVM.

**A.2. Advanced Deep Learning Techniques:** Advanced methods involving two-tier transfer learning were utilized in studies which used architectures like MobileNet, InceptionV2, and ResNet101. Goyal et al. [32] presented DFUNet for classification of DFUs, while Nilsson et al. [33] applied a CNN-based method using VGG-19 for venous ulcer image classification. In another significant study, Alaskar et al. [38] applied deep CNNs for intestinal ulcer detection in wireless capsule endoscopy images. Using AlexNet and GoogleNet architectures, they reported a classification accuracy of 100% for both networks.
Ahsan et al. [41] discusses the use of deep learning algorithms to automatically classify diabetic foot ulcers (DFU), a serious complication of diabetes that can lead to lower limb amputation if untreated. The authors examined various convolutional neural network (CNN) architectures, including AlexNet, VGG16/19, GoogleNet, ResNet50, MobileNet, SqueezeNet, and DenseNet. They used these models to categorize infection and ischemia in the DFU2020 dataset. To address the issue of limited data and to reduce computational cost, they fine-tuned the weights of the models. Additionally, affine transform techniques were employed for data augmentation. The results revealed that the ResNet50 model achieved the highest accuracy rates, reaching 99.49% for ischemia and 84.76% for infection detection.

**A.3. Multi-Class Classification Techniques:** Shenoy et al. [34] proposed a method to classify wound images into multiple classes using deep CNNs. Rostami et al. [36] proposed an ensemble DCNN-based classifier to classify entire wound images into surgical, diabetic, and venous ulcers. Anisuzzaman et al. [28] proposed a multi-modal classifier using wound images and their corresponding locations to categorize them into multiple classes, including diabetic, pressure, surgical, and venous ulcers. This paper introduced an image and location classifier and combined it together to create a multi-modal classifier. In this study, two different datasets were used namely AZH dataset that consists of 730 wound images with four classes, Medetec dataset which consists of 358 wound images with three classes. Also, they introduced a new dataset AZHMT dataset which is a combination of AZH and Medetec dataset containing 1088 wound images. The reported maximum accuracy on mixed-class classifications varies from 82.48% to 100% in different experiments and maximum accuracy on wound-class classifications varies from 72.95% to 97.12% in various experiments.

### B. Wound Image Classification Using Novel Approaches

Novel techniques have been presented to overcome challenges in wound classification. Alzubaidi et al. [35] introduced DFU_QUTNet, a deep architecture model for classification of DFUs. They reported a maximum F1-Score of 94.5% obtained from combining DFU_QUTNet and SVM. Another interesting method was presented by Sarp et al. [37] who classified chronic wounds using an explainable artificial intelligence (XIA) approach.

### C. Traditional Machine Learning-Based Classification

**C.1. SVM-Based Techniques:** Traditional machine learning techniques have also found significant use in wound image classification. Yadav et al. [39] used color-based feature extraction and SVM for binary classification of burn wound images. Goyal et al. [40] used traditional machine learning and DCNN techniques for detecting and localizing DFUs, with Quadratic SVM classifiers trained on feature-rich patches extracted from the images.

Through this review, it is evident that both traditional and advanced machine learning techniques have demonstrated promising results in wound image classification, providing valuable insights for future research in this area.

## Materials and Methods

This study encompasses three distinct subsections, each elucidating the specific methodology employed in this study: Whole Image Classification, Region of Interest (ROI) Extracted Image Classification, and ROI with Body Map Location Image Classification. It should be noted that each of these subsections utilizes the same fundamental base classifier for the image data analysis. Datasets were anonymized, partitioned, and augmented before processing through a proposed architecture. The proposed model incorporated transfer learning, convolution blocks, axial-attention mechanisms, and Adaptive-gated MLP. Model performance was evaluated using accuracy, precision, recall, and the F1-score.

### A. Dataset

**A.1 AZH Dataset:** The AZH Dataset is a collection of prefiltered 730 ROI images and 538 Whole wound images, varying in size and depicting four types of wounds: venous, diabetic, pressure, and surgical. Captured over two years at Milwaukee's AZH Wound and Vascular Center, the images were collected using an iPad Pro (software version 13.4.1) and a Canon SX 620 HS digital camera, and subsequently labeled by a wound specialist from the center. While most of the dataset comprises unique patient cases, some instances involve multiple images from a single patient, taken from different body sites or at varying stages of healing. These were classified as separate due to distinct wound shapes. This dataset, unfortunately, couldn't be expanded due to resource limitations. It's important to note that the data doesn't involve any human experimentation or usage of human tissue samples. Instead, it utilizes de-identified wound images, publicly available at link:

(Link). Each image only includes the wound and immediate skin area, eliminating any unnecessary or personal data to protect patient identity. The University of Wisconsin-Milwaukee has vetted the dataset's use for compliance with university policy. **Figure 3** and **Figure 4** show images from whole and ROI images.

**A.2 Medetec Dataset:** The Medetec wound dataset is a compendium of freely available images that encompasses an extensive range of open wounds [57]. We prefiltered 216 images from three distinct categories for this study: diabetic wounds, pressure ulcers, and venous leg ulcers. Notably, this dataset does not encompass images of surgical wounds. The images are provided in .jpg format, with weights and heights fluctuating between 358 and 560 pixels, and 371 to 560 pixels, respectively. This dataset laid a solid foundation for the robustness and reliability assessments of the model we developed.

**B. Body map for location**

A body map serves as a simplified, symbolic, and accurately phenotypic representation of an individual's body [42]. Primarily used in the medical field, body maps are effective tools for identifying and locating physical afflictions such as bruises, wounds, or fractures. They are especially valuable in forensic science for identifying bodily changes during post-mortem examinations and in medical practice for pinpointing the location of infections [43]. By offering a detailed overview of the body, they inform practitioners about other body areas that might be affected and require attention during the healing process. Furthermore, in the realm of scientific research, body maps function as verifiable evidence, validating observable bodily changes caused by internal diseases.

The design of a comprehensive body map with 484 distinct parts is credited to Anisuzzaman et al. [28]. PaintCode [44] was employed to prepare this body map, with initial references being drawn from several credible sources [45][46][47]. The fundamental framework for this design originated from the Original Anatomy Mapper [48], which directly paired each label and outline. The extreme intricacy involved in the detailed depiction of each feature on the body map led to a pre-selection of 484 features or regions. This process was overseen and approved by wound professionals at the AZH wound and vascular center, ensuring the map's medical accuracy and applicability. Major part of body map is shown in the **Figure *2***. Each number denotes a location in this case. **Table 1** shows a few examples of locations and their related numbers.

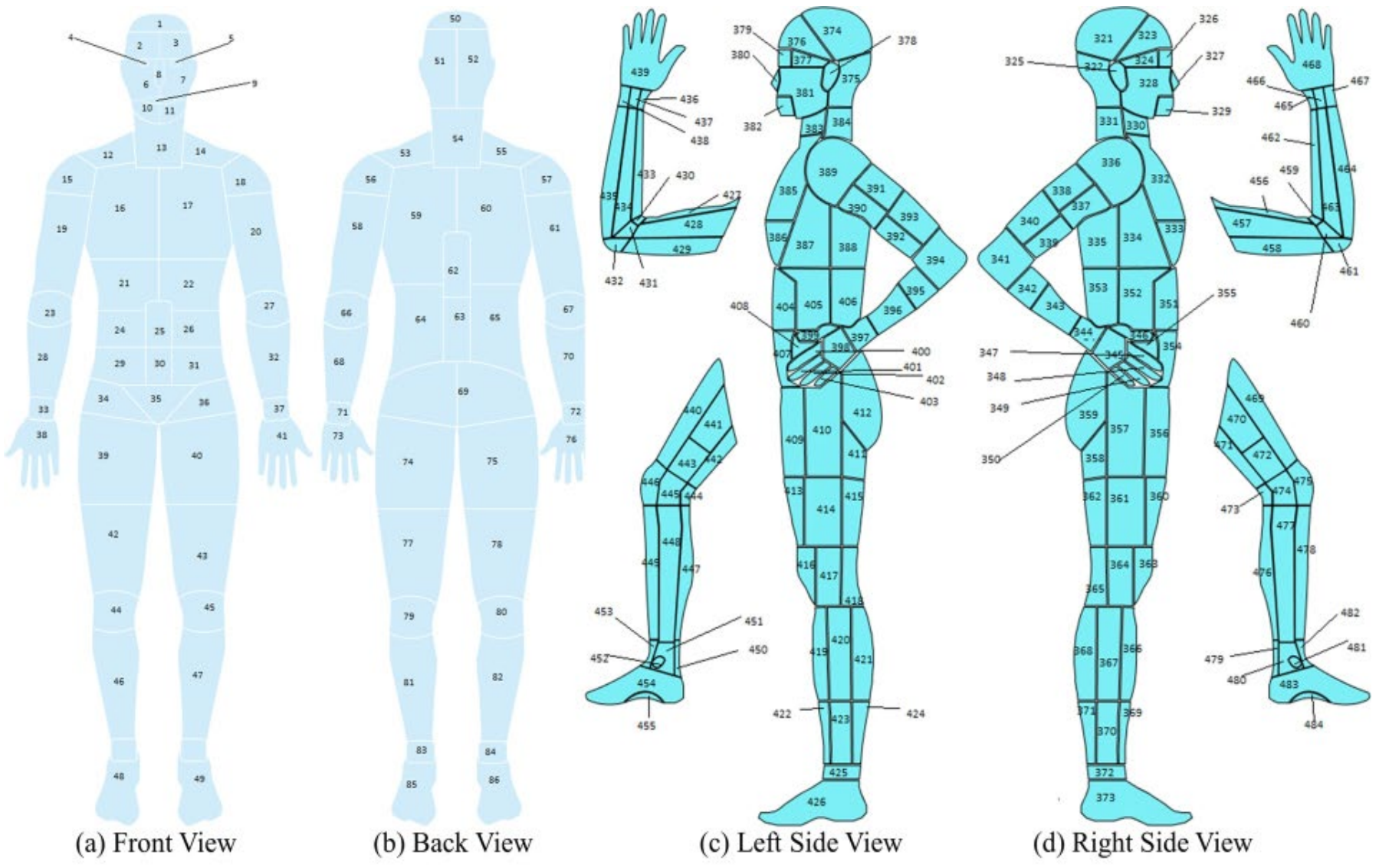

**Figure 2:** Full Body View [28]

| Right Leg Front and Back | | Left Leg Front and Back | |
|---|---|---|---|
| **Location name** | **Body map number** | **Location name** | **Body map number** |
| Right Fifth Toe Tip | 135 | Left Anterior Ankle | 180 |
| Right Lateral Heel | 150 | Left Fifth Toe Tip | 202 |
| Right Medial Malleolus | 158 | Left Medial Malleolus | 178 |
| Right Proximal Lateral Dorsal Foot | 159 | Left Proximal Medial Plantar Foot | 215 |

**Table 1**: Body Map examples of Lower leg region

## C. Dataset Processing and Augmentation

**C.1. ROI extraction:** The extraction of Region of Interest (ROI) from wound images presents a robust methodology for diagnosing and tracking wound progression. As aforementioned, the ROI includes the wound itself and a portion of the surrounding healthy skin, which collectively encompasses the vital elements of the wound's condition. The developed wound localizer is a significant tool for this extraction process, as it is capable of automatically cropping single or multiple ROIs from each image [49].

Each ROI represents one of six categories - diabetic, venous, pressure, surgical, background, and normal skin. These categories are critical in understanding the etiology of the wound, allowing for more accurate and personalized treatment plans. However, the diversity of the wounds is also reflected in the different sizes and shapes of the extracted ROIs, each telling a unique narrative of the wound's journey.

Importantly, the ROI's rectangular form and variable size allow for greater adaptability in handling various wound types and sizes. It is an efficient method to focus on the essential wound characteristics while reducing unnecessary information that could potentially introduce noise into the data.

**Figure 4** excellently illustrates the variation in extracted ROIs from different classes of the wound dataset. This showcases the versatility of our wound localizer, capable of handling wounds of different origins, sizes, and stages. It successfully extracts the ROI, making the most relevant information available for analysis.

**C.2. Data Split:** During this study, we utilize two distinct methods to partition the dataset. The utilization of two distinct partitioning provides a richer understanding of the model's behavior, potential biases, sensitivities, and the ability to generalize. It also facilitates a more robust and comprehensive evaluation process. The first approach consisted of splitting the data into training (70%), testing (15%), and validation (15%). The second partitioning method diverged slightly from the first, allocating 60% of the data for training, 15% for validation, and increasing the testing set to comprise 25% of the total dataset. **Table 2** shows both types of dataset splits on ROI images.

**C.3. Data Augmentation:** Each image in the training set was augmented using transformation methods such as resizing, rotation, flipping (both vertically and horizontally), and application of affine transforms (including scaling, rotating, and shifting). Additional alterations such as the application of Gaussian noise and coarse dropout (i.e., random rectangular region removal) were also performed. These transformations were probabilistically applied, creating a diverse set of augmented samples as shown in **Figure 5**. The transformations ensured robustness of the model against variations in the data.

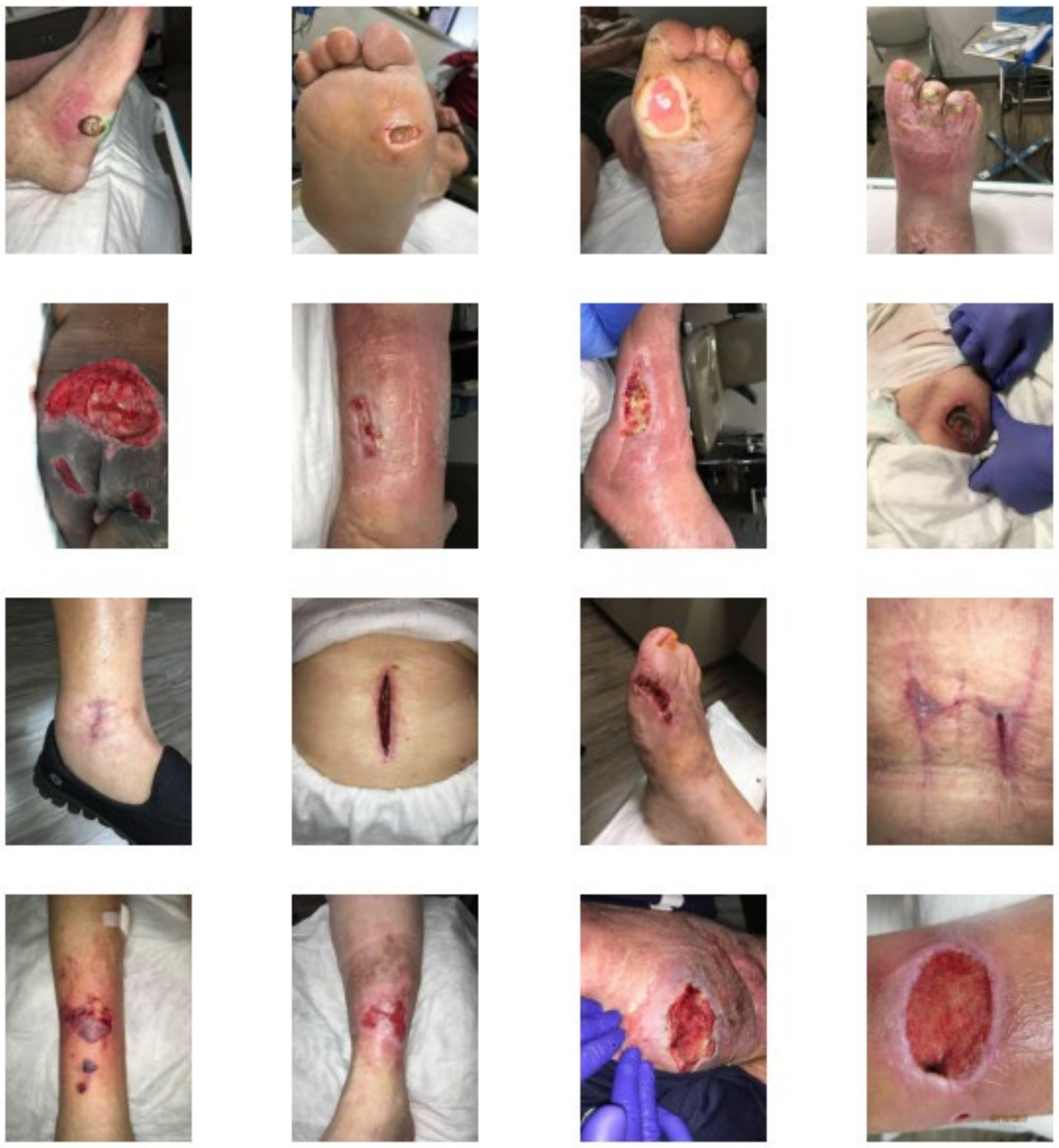

**Figure 3:** Sample images from the AZH Wound and Vascular Center database. The rows from top to bottom display diabetic, pressure, surgical and venous samples, respectively.

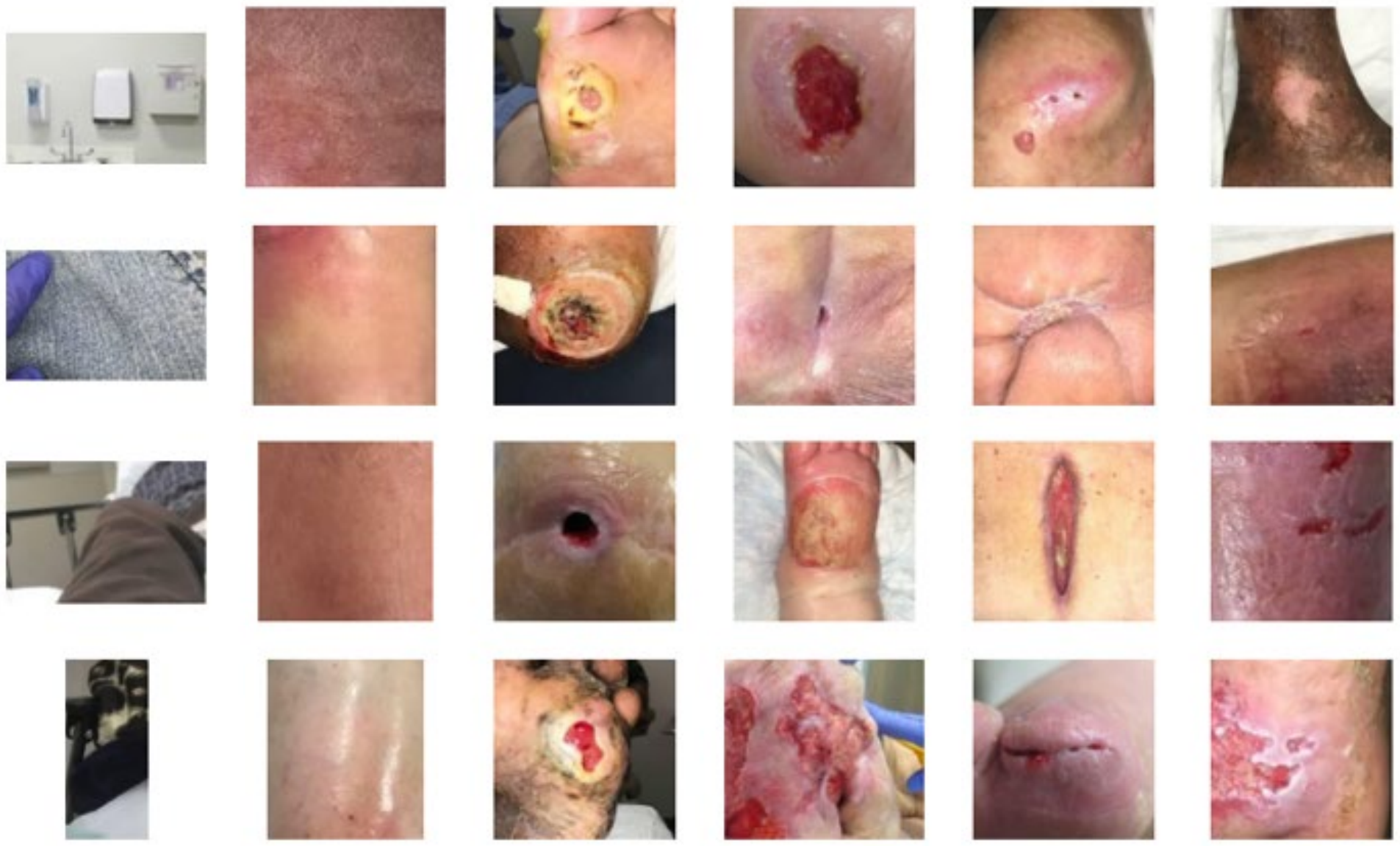

**Figure 4:** Sample ROIs. The columns from left to right display background, normal skin, diabetic, pressure, surgical and venous ROIs, respectively.

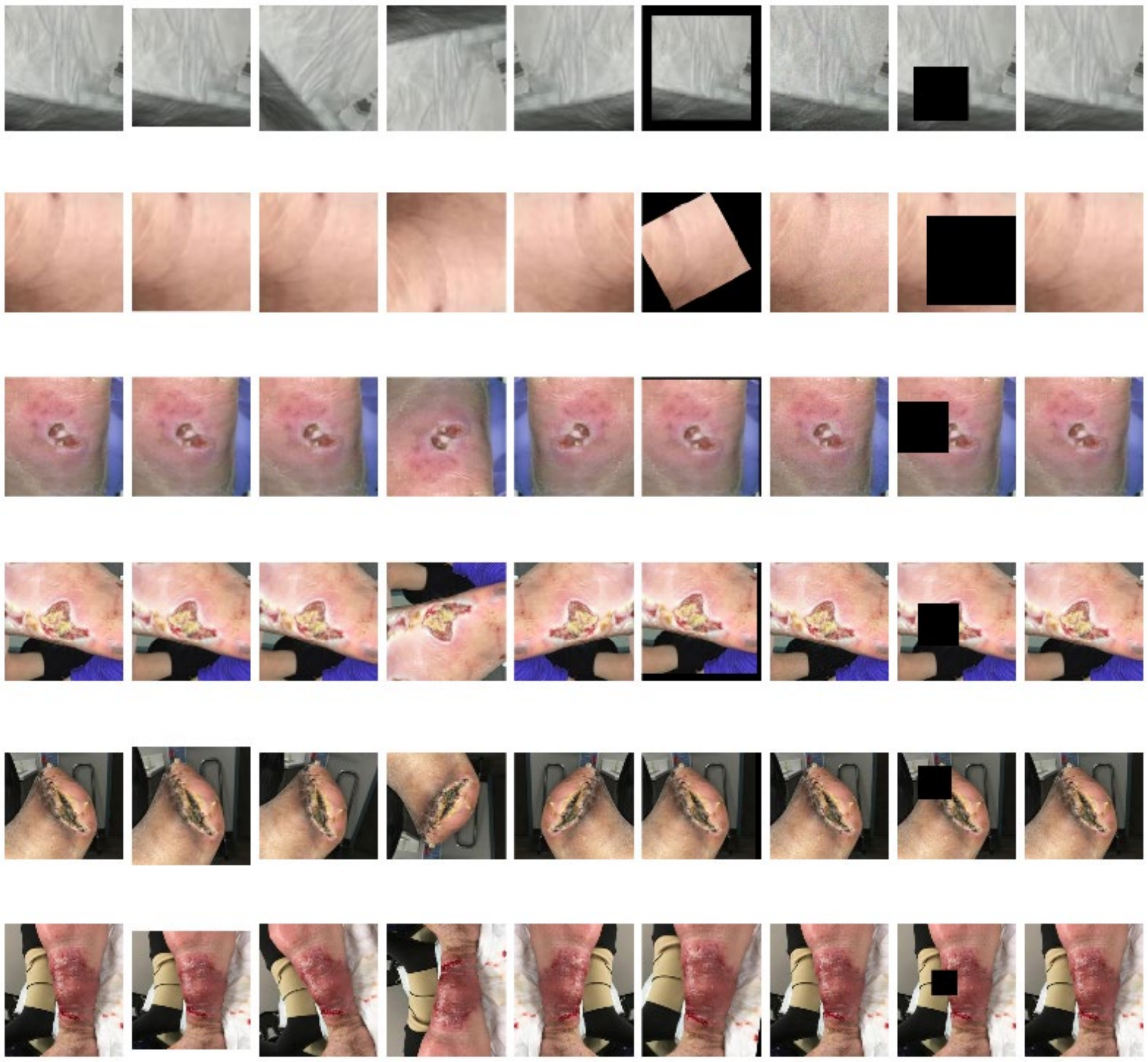

**Figure 5**: Data Augmentation with leftmost original image. The rows from top to bottom display background, normal skin, diabetic, pressure, surgical and venous ROIs, respectively.

| Class Abbreviation | | Back ground | Normal Skin | Venous | Diabetic | Pressure | Surgical | Sub Total | Total |
|---|---|---|---|---|---|---|---|---|---|
| | | BG | N | V | D | P | S | | |
| **ROI image-based Dataset: 60% training, 15% validation, 25% testing** | | | | | | | | | |
| **AZH** | Training | 60 | 60 | 148 | 111 | 80 | 98 | 557 | 930 |
| | Validation | 15 | 15 | 37 | 27 | 20 | 24 | 138 | |
| | Testing | 25 | 25 | 62 | 47 | 34 | 42 | 235 | |
| **ROI image-based Dataset: 70% training, 15% validation, 15% testing** | | | | | | | | | |
| **AZH** | Training | 70 | 70 | 172 | 129 | 93 | 114 | 648 | 930 |
| | Validation | 15 | 15 | 37 | 27 | 20 | 24 | 138 | |
| | Testing | 15 | 15 | 38 | 29 | 21 | 26 | 144 | |
| **Whole image-based Dataset: 60% training, 15% validation, 25% testing** | | | | | | | | | |
| **AZH** | Training | - | - | 93 | 92 | 60 | 76 | 321 | 538 |
| | Validation | - | - | 23 | 23 | 15 | 19 | 80 | |
| | Testing | - | - | 40 | 39 | 25 | 33 | 137 | |
| **Whole image-based Dataset: 70% training, 15% validation, 15% testing** | | | | | | | | | |
| **AZH** | Training | - | - | 109 | 107 | 70 | 89 | 375 | 538 |
| | Validation | - | - | 23 | 23 | 15 | 19 | 80 | |
| | Testing | - | - | 24 | 24 | 15 | 20 | 83 | |
| **Whole image-based Dataset: 60% training, 15% validation, 25% testing** | | | | | | | | | |
| **Medetec** | Training | - | - | 37 | 27 | 65 | - | 129 | 216 |
| | Validation | - | - | 9 | 6 | 16 | - | 31 | |
| | Testing | - | - | 16 | 12 | 28 | - | 56 | |
| **Whole image-based Dataset: 70% training, 15% validation, 15% testing** | | | | | | | | | |
| **Medetec** | Training | - | - | 43 | 31 | 76 | - | 150 | 216 |
| | Validation | - | - | 9 | 6 | 16 | - | 31 | |
| | Testing | - | - | 10 | 8 | 17 | - | 35 | |

**Table 2**: AZH and Medetec original dataset count and class Abbreviations

**C.4. ROI and wound location:** In the ROI dataset we have two additional classes named normal skin and background which were created manually by selecting skin region for normal skin and any additional information as background from the original whole image dataset **Error! Bookmark not defined.**[28]. Sample of these two classes are shown in **Figure 4**. All of these were verified by wound specialists. Wound location was associated with each ROI image and assigned values from the body map discussed in section B. All the six classes abbreviation is shown in **Table 2.**

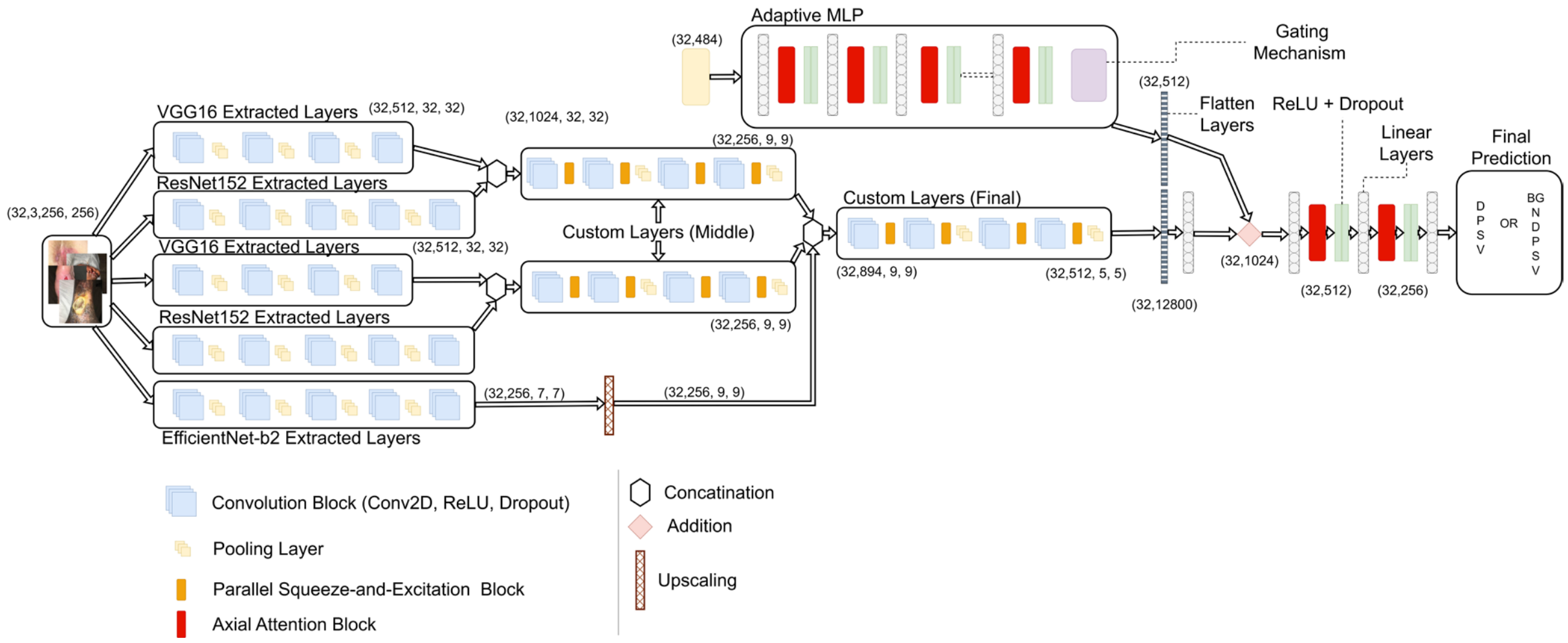

**Figure 6**: Proposed model architecture outline

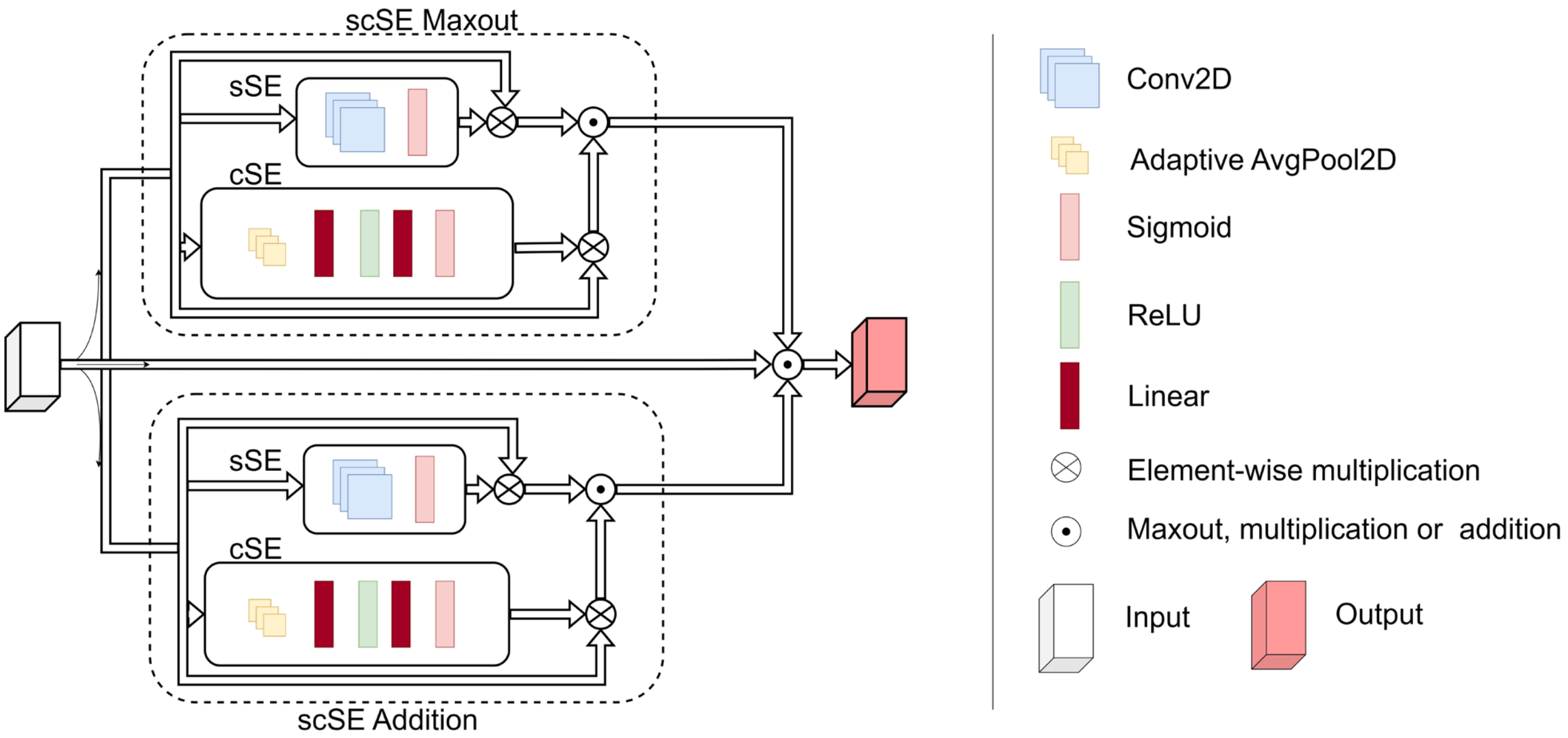

**Figure 7:** Parallel Squeeze-and-Excitation block architecture outline

## D. Model

Our proposed deep learning model integrates multi-level features from various pre-existing models, utilizing custom layers and attention mechanisms to improve performance. Our model design has been adopted from C-Net architecture [53][54]. Basic model outline is displayed in **Figure 6**.

**D.1 Base Models:** The proposed model utilizes three pre-trained Convolutional Neural Networks (CNNs) - ResNet152, VGG16, and EfficientNet-B2. In ResNet152, modifications include not only the removal of the average pooling and fully connected layers, but also alterations to the third block of the second layer by removing its last two layers and removing the final four layers of the overall model. For VGG16, the last twelve layers are omitted, capturing more primitive patterns. The last layer of EfficientNet-B2 is removed to maintain consistency with the modifications made to the other two models. These models, applied in parallel to the input, capture different levels of features.

**D.2 Custom Layers:** The custom layers comprise a Convolutional Block (ConvBlock), followed by a Parallel Squeeze-and-Excitation (P_scSE) block [51], and a dropout layer. The ConvBlock is a combination of a convolution layer and a ReLU activation function, capturing spatial information and introducing non-linearity.

The P_scSE block blends Channel-wise Squeeze-and-Excitation (cSE) and Spatial Squeeze-and-Excitation (sSE) operations. The cSE focuses on channel interdependencies, providing global context, while the sSE concentrates on the spatial interdependencies of each channel, maintaining detailed spatial information. Outputs from the cSE and sSE are merged using max-out and addition operations [51] as shown in **Figure 7**.

**D.3 Aggregation and Fully Connected Layers:** The base models' outputs are concatenated and fed through sequences of ConvBlocks and P_scSE blocks to merge and process multi-level features. The output is then flattened and passed through a dense layer.

The output is further processed through a fully connected layer block. This block includes two dense layers enriched with axial-attention mechanisms, an enhancement over traditional attention mechanisms, focusing on individual dimensions separately. Interspersed with ReLU activation functions and dropout operations, the axial-attention mechanism boosts important features, helping the model to recognize complex patterns and dependencies in the data.

If wound location is used along with image data, then we use Adaptive-gated MLP to analyze wound location separately before concatenating them with the fully connected layers in the above model. This module is constructed as a series of linear transformations, axial attentions, and ReLU activations, followed by a final linear transformation to map the output to the target size. The MLP is gated, meaning it can learn to selectively propagate information through the network.

The orderly arrangement of data is vital for the efficient functioning of the model. Consistency in the output from the image and location data is essential, thus necessitating the synchronous feeding of properly sequenced data into the model. This alignment was maintained by associating each Region of Interest (ROI) with a unique index number and mapping the corresponding wound location to this number. Given the categorical nature of wound location data, it was represented using one-hot encoding.

**D.4 Output Layer:** The final dense layer maps to the number of output classes.

**E. Performance Metrics**

In our study, we employed various evaluation metrics such as accuracy, precision, recall, and the F1-score to scrutinize the effectiveness of the classifiers. The related mathematical formulations for these assessment metrics are illustrated in Equations 1 through 4. The abbreviations TP, TN,

FP, and FN in these equations stand for True Positive, True Negative, False Positive, and False Negative, respectively. For a more comprehensive understanding of these equations and associated theories, readers are referred to reference [50].

$$Accuracy = \frac{TP + TN}{TP + FP + FN + TN} \tag{1}$$

$$Precision = \frac{TP}{TP + FP} \tag{2}$$

$$Recall = \frac{TP}{TP + FN} \tag{3}$$

$$F1 - Score = 2 \times \frac{Recall \times Precision}{Recall + Precision} \tag{4}$$

## Results

In the present investigation, we deployed the advanced computational capacities of Google Colab Pro Plus A100, fortified with 40GB of memory. This enabled a methodical analysis involving both Region of Interest (ROI) and whole image-based classifications. The experimental setup involved processing images of 256x256 pixel dimensions, batched into groups of 32, across a course of 100 epochs. Our learning parameters were finely tuned to optimize the learning process: a learning rate of 0.0001 was chosen, with a minimum rate limit of 0.00001. To enhance the efficiency of our learning process, we applied the Adam optimizer [52].

**Classification Categories:** The classifiers were extensively trained to distinguish among various classes represented in the images, specifically: Diabetic (D), Venous (V), Pressure (P), Surgical (S), Background (BG), and Normal Skin (N). Further specifications and results regarding these classes will be provided in the ensuing sections of this paper.

**Loss Function:** Cross Entropy was chosen as our loss function, given the multi-class and binary nature of our image classifications. Its mathematical formulation is as follows 26[56]:
For multi-class problems, the cross-entropy loss, $L$, is:

$$L = -\sum_{i=0}^{n} (y_i \times \log{(p_i)}) \tag{1}$$

Here $y_i$ is the actual label and ($p_i$) is the predicted probability for each class ($i$). For binary classification problem, the binary cross entropy loss, $L$, is computed as:

$$L = -\sum_{i=0}^{n} (y_i \times \log(p_i) + (1 - y_i) \times \log(1 - p_i)) \tag{2}$$

The optimization process strives to minimize this loss, thereby reducing the discrepancy between our model's predictions ($p_i$) and the actual labels ($y_i$). Further sections will elucidate the efficacy of this loss function within our research context.

## A. ROI Classification

The primary phase of the ROI classification trial pertains to the classification of 6 unique types of wound patches, specifically: diabetic, venous, pressure, surgical, BG, and N. Subsequently, the 5-category classification problem comprised three types of wound labels alongside BG and N categories. When addressing the 4-category classification, the objective centered on the categorization of the wound patches into one of the four classes: BG, N, along with two different wound labels. In the context of 3-category classification, the aim was to sort the wound patches into one of the three groups: D, P, S, V. For binary classification, a range of combinations including N, D, P, S, V were utilized to categorize the wound patches into two distinct groups. The dataset was split in two different ways, one is 70-15-15 and the other is 60-15-25, to observe and compare the best results.

**A.1. ROI multiclass classification without wound location**: The results of the ROI classifier's performance without wound location evaluation varied across different scenarios. For the 6-class classification case (BG, N, D, P, S, V), the test accuracy was 85.41% and 80.42% for the 70%, 15%, 15% and 60%, 15%, 25% data splits respectively. The precision, recall and F1-score for this case were 85.69%, 85.41%, 85.29% and 80.26%, 80.42%, 79.52% for each data split respectively, as displayed in **Table 3**.

In the 5-class classification scenario, the results varied between the class combinations. The BG, N, D, S, V combination showed superior performance with test accuracies, precisions, recalls, and F1-scores of 91.86%, 92.29%, 91.86%, 91.91% and 91.04%, 91.30%, 91.04%, 90.96% for each data split respectively. Conversely, the BG, N, D, P, S class combination registered slightly lower accuracy rates of 87.73% and 84.39%, along with precision, recall and F1-score values of 88.91%, 87.73%, 87.74% and 84.39%, 84.39%, 84.39% for each data split respectively.

When the classifier was tested for 4-class classification, BG, N, D, V demonstrated high accuracy rates of 96.90% and 96.22%, with precision, recall, and F1-score of 97.04%, 96.90%, 96.90% and 96.31%, 96.22%, 96.23% for each data split respectively. However, the BG, N, P, S combination indicated a decrease in accuracy at 87.01% and 85.71%, along with precision, recall, and F1-score values of 89.16%, 87.01%, 87.30% and 85.88%, 85.71%, 85.78% for each data split respectively.

The performance for 3-class and 2-class classification showed a range of accuracy scores, with the 2-class case achieving 100% accuracy for the N, D combination in both data splits, with corresponding precision, recall, and F1-score values also being 100%. All these experiments were performed with data augmentation only on train data, as it consistently led to improved results.

| No. of Classes | Classes | A | P | R | F |
|---|---|---|---|---|---|
| **6 Class** | **BG, N, D, P, S, V** | **85.41** | 85.69 | 85.41 | 85.29 |
| **5 Class** | **BG, N, D, P, V** | 88.13 | 89.01 | 88.13 | 87.80 |
| | **BG, N, D, S, V** | **91.86** | 92.29 | 91.86 | 91.91 |
| | **BG, N, D, P, S** | 87.73 | 88.91 | 87.73 | 87.74 |
| | **BG, N, P, S, V** | 86.95 | 86.83 | 86.95 | 86.56 |
| **4 Class** | **BG, N, D, V** | **96.90** | 97.04 | 96.90 | 96.90 |
| | **BG, N, P, V** | 95.50 | 95.57 | 95.50 | 95.48 |
| | **BG, N, S, V** | 94.68 | 94.76 | 94.68 | 94.69 |
| | **BG, N, D, P** | 87.50 | 87.58 | 87.50 | 87.47 |
| | **BG, N, D, S** | 90.58 | 91.20 | 90.58 | 90.59 |
| | **BG, N, P, S** | 87.01 | 89.16 | 87.01 | 87.30 |
| **3 Class** | **D, S, V** | **91.39** | 91.93 | 91.39 | 91.30 |
| | **P, S, V** | 87.05 | 87.18 | 87.05 | 87.10 |
| | **D, P, S** | 82.89 | 83.64 | 82.89 | 82.41 |
| | **D, P, V** | 86.36 | 85.99 | 86.36 | 85.75 |
| **2 Class** | **N, D** | **100.0** | 100.0 | 100.0 | 100.0 |
| | **N, P** | 94.44 | 95.09 | 94.44 | 94.47 |
| | **N, S** | **100.0** | 100.0 | 100.0 | 100.0 |
| | **N, V** | 98.11 | 98.23 | 98.11 | 98.13 |
| | **D, P** | 88.00 | 88.00 | 88.00 | 88.00 |
| | **D, S** | 90.90 | 91.35 | 90.90 | 90.85 |
| | **D, V** | 98.50 | 98.55 | 98.50 | 98.51 |
| | **P, S** | 85.10 | 85.23 | 85.10 | 85.13 |
| | **P, V** | 93.22 | 93.29 | 93.22 | 93.13 |
| | **S, V** | 93.75 | 93.85 | 93.75 | 93.70 |

| No. of Classes | Classes | A | P | R | F |
|---|---|---|---|---|---|
| **6 Class** | **BG, N, D, P, S, V** | **80.42** | 80.26 | 80.42 | 79.52 |
| **5 Class** | **BG, N, D, P, V** | **83.41** | 83.34 | 83.41 | 83.20 |
| | **BG, N, D, S, V** | 91.04 | 91.30 | 91.04 | 90.96 |
| | **BG, N, D, P, S** | 84.39 | 84.39 | 84.39 | 84.39 |
| | **BG, N, P, S, V** | 87.23 | 87.25 | 87.23 | 87.15 |
| **4 Class** | **BG, N, D, V** | **96.22** | 96.31 | 96.22 | 96.23 |
| | **BG, N, P, V** | 91.09 | 91.21 | 91.09 | 90.96 |
| | **BG, N, S, V** | 91.55 | 91.68 | 91.55 | 91.51 |
| | **BG, N, D, P** | 87.02 | 87.43 | 87.02 | 87.16 |
| | **BG, N, D, S** | 89.2 | 91.81 | 89.20 | 89.06 |
| | **BG, N, P, S** | 85.71 | 85.88 | 85.71 | 85.78 |
| **3 Class** | **D, S, V** | **90.72** | 90.98 | 90.72 | 90.49 |
| | **P, S, V** | 83.33 | 83.26 | 83.33 | 83.01 |
| | **D, P, S** | 74.79 | 74.28 | 74.79 | 74.39 |
| | **D, P, V** | 85.31 | 85.20 | 85.31 | 84.71 |
| **2 Class** | **N, D** | **100.0** | 100.0 | 100.0 | 100.0 |
| | **N, P** | 96.61 | 96.61 | 96.61 | 96.61 |
| | **N, S** | 98.5 | 98.54 | 98.50 | 98.50 |
| | **N, V** | **100.0** | 100.0 | 100.0 | 100.0 |
| | **D, P** | 85.18 | 87.05 | 85.18 | 84.62 |
| | **D, S** | 89.88 | 90.34 | 89.88 | 89.82 |
| | **D, V** | 97.24 | 97.27 | 97.24 | 97.25 |
| | **P, S** | 81.57 | 81.56 | 81.57 | 81.51 |
| | **P, V** | 90.62 | 90.61 | 90.62 | 90.51 |
| | **S, V** | 93.26 | 93.25 | 93.26 | 93.25 |

**Table 3**: ROI image without location-based classification with different data split (Left - 70%,15%,15%, Right - 60%,15%,25%). P = Precision, R = Recall, F = F1-score, Acc = Accuracy

**A.2. ROI multi-class classification with wound location**: Following the inclusion of wound location data in conjunction with image data, **Table 4** displays the performance metrics from experiments using an Adaptive-gated MLP to separately analyze the wound location. This data was subsequently concatenated with the fully connected layers of the prior model.

For the 6-class classification comprising BG, N, D, P, S, and V classes, the accuracy was recorded at 87.50% and 83.82%, precision at 88.04% and 83.42%, recall at 87.50% and 83.82%, and F1-score at 87.37% and 83.53% for the data splits of 70%,15%,15% and 60%,15%,25% respectively. Moving on to the 5-class classification, the class combination BG, N, D, S, V saw strong results with accuracy levels of 91.86% and 91.54%, precision at 91.99% and 91.65%, recall at 91.86% and 91.54%, and F1-score at 91.85% and 91.50% across the two data splits. Conversely, the BG, N, D, P, S combination demonstrated lower accuracy at 84.90% and 84.39%, precision at 85.28% and 85.56%, recall at 84.90% and 84.39%, and F1-score at 84.96% and 83.92%.

In the context of the 4-class classification, the BG, N, D, V combination once again showed impressive metrics with accuracy rates of 95.87% and 96.22%, precision at 96.06% and 96.37%, recall at 95.87% and 96.22%, and F1-score at 95.83% and 96.24%. On the other hand, the BG, N, P, S combination witnessed a decrease in performance, registering accuracy levels of 90.90% and 88.88%, precision at 91.50% and 88.90%, recall at 90.90% and 88.88%, and F1-score at 91.03% and 88.72% for each respective data split.

For the 3-class and 2-class classification models, a range of performance scores were observed. The 2-class case, particularly the N, D combination, achieved perfect performance with accuracy, precision, recall, and F1-score all at 100% in both data splits. The D, P class combination, however, recorded the lowest performance levels for this category with accuracy at 86.14% and 86.41%, precision at 86.14% and 86.73%, recall at 86.00% and 86.41%, and F1-score at 86.03% and 86.22%.

In conclusion, the results show that the incorporation of wound location data alongside image data led to variations in accuracy, precision, recall, and F1-score based on the number and combination of classes, as well as the distribution of the data split. Furthermore, the use of an Adaptive-gated MLP for separate wound location analysis consistently resulted in promising outcomes across all experiments.

| No. of Classes | Classes | A | P | R | F |
|---|---|---|---|---|---|
| 6 Class | BG, N, D, P, S, V | **87.50** | 88.04 | 87.50 | 87.37 |
| 5 Class | BG, N, D, P, V | 91.52 | 91.51 | 91.52 | 91.44 |
| | BG, N, D, S, V | **91.86** | 91.99 | 91.86 | 91.85 |
| | BG, N, D, P, S | 84.90 | 85.28 | 84.90 | 84.96 |
| | BG, N, P, S, V | 86.70 | 86.99 | 86.70 | 86.34 |
| 4 Class | BG, N, D, V | 95.87 | 96.06 | 95.87 | 95.83 |
| | BG, N, P, V | 94.38 | 94.50 | 94.38 | 94.34 |
| | BG, N, S, V | **96.80** | 97.04 | 96.80 | 96.80 |
| | BG, N, D, P | 88.75 | 89.05 | 88.75 | 88.78 |
| | BG, N, D, S | 92.94 | 93.24 | 92.94 | 92.74 |
| | BG, N, P, S | 90.90 | 91.50 | 90.90 | 91.03 |
| 3 Class | D, S, V | **92.47** | 92.79 | 92.47 | 92.34 |
| | P, S, V | 87.05 | 87.18 | 87.05 | 87.10 |
| | D, P, S | 81.57 | 81.69 | 81.57 | 81.07 |

| No. of Classes | Classes | A | P | R | F |
|---|---|---|---|---|---|
| 6 Class | BG, N, D, P, S, V | 83.82 | 83.42 | 83.82 | 83.53 |
| 5 Class | BG, N, D, P, V | 89.11 | 88.88 | 89.11 | 88.71 |
| | BG, N, D, S, V | **91.54** | 91.65 | 91.54 | 91.50 |
| | BG, N, D, P, S | 84.39 | 85.56 | 84.39 | 83.92 |
| | BG, N, P, S, V | 88.82 | 88.81 | 88.82 | 88.70 |
| 4 Class | BG, N, D, V | **96.22** | 96.37 | 96.22 | 96.24 |
| | BG, N, P, V | 93.15 | 93.52 | 93.15 | 93.10 |
| | BG, N, S, V | **96.10** | 96.19 | 96.10 | 96.05 |
| | BG, N, D, P | 89.31 | 90.50 | 89.31 | 89.06 |
| | BG, N, D, S | 93.52 | 93.72 | 93.52 | 93.57 |
| | BG, N, P, S | 88.88 | 88.90 | 88.88 | 88.72 |
| 3 Class | D, S, V | **90.72** | 91.23 | 90.72 | 90.64 |
| | P, S, V | 84.05 | 83.87 | 84.05 | 83.38 |
| | D, P, S | 73.98 | 76.52 | 73.98 | 71.58 |

| | | | | | | | | | | | |
|---|---|---|---|---|---|---|---|---|---|---|---|
| | **D, P, V** | 89.77 | 90.04 | 89.77 | 89.58 | | **D, P, V** | 86.71 | 86.53 | 86.71 | 86.47 |
| **2 Class** | **N, D** | **100.0** | 100.0 | 100.0 | 100.0 | **2 Class** | **N, D** | **100.0** | 100.0 | 100.0 | 100.00 |
| | **N, P** | 94.44 | 95.09 | 94.44 | 94.47 | | **N, P** | 96.61 | 96.61 | 96.61 | 96.61 |
| | **N, S** | 97.56 | 97.65 | 97.56 | 97.54 | | **N, S** | 98.50 | 98.54 | 98.50 | 98.50 |
| | **N, V** | 98.11 | 98.16 | 98.11 | 98.09 | | **N, V** | 98.85 | 98.89 | 98.85 | 98.85 |
| | **D, P** | 86.14 | 86.14 | 86.00 | 86.03 | | **D, P** | 86.41 | 86.73 | 86.41 | 86.22 |
| | **D, S** | 92.72 | 92.97 | 92.72 | 92.70 | | **D, S** | 89.88 | 90.34 | 89.88 | 89.82 |
| | **D, V** | 98.50 | 98.55 | 98.50 | 98.51 | | **D, V** | 97.24 | 97.27 | 97.24 | 97.25 |
| | **P, S** | 87.23 | 87.61 | 87.23 | 87.26 | | **P, S** | 84.21 | 84.39 | 84.21 | 84.24 |
| | **P, V** | 96.61 | 96.77 | 96.61 | 96.56 | | **P, V** | 92.70 | 92.99 | 92.70 | 92.56 |
| | **S, V** | 95.31 | 95.65 | 95.31 | 95.25 | | **S, V** | 94.23 | 94.42 | 94.23 | 94.17 |

**Table 4**: ROI image with location-based classification with different data split (Left - 70%,15%,15%, Right - 60%,15%,25%). P = Precision, R = Recall, F = F1-score, A = Accuracy

## B. Whole Image Classification

In the whole image classification, the precision, recall, and F1-score measurements show that the incorporation of these metrics, alongside accuracy, provides a more comprehensive understanding of the model's performance. **Table 5** depicts these additional measurements, and they reveal interesting patterns that match with the observed accuracy rates.

For the 4-class classification comprising D, P, S, and V, precision, recall, and F1-scores were observed at 83.22%, 83.13%, and 82.26% respectively for the 70-15-15 data split. For the 60-15-25 split, these scores were slightly lower, coming in at 78.60%, 78.10%, and 76.75%, respectively. This pattern is similarly reflected in the accuracy measurements for the same class combination and data splits.

In the 3-class classification, the D, S, V combination showed a high precision of 93.48%, recall of 92.64%, and F1-score of 92.54% for the 70-15-15 split. Conversely, the D, P, S combination demonstrated lower values, with a precision of 82.66%, recall of 81.35%, and F1-score of 80.72% in the same split.

Focusing on the 2-class classification, all N-related combinations (N, D; N, P; N, S; N, V) achieved perfect precision, recall, and F1-score of 100% in both data splits. However, other combinations like D, P and P, S displayed lower scores. The D, P combination, for instance, recorded precision, recall, and F1-score of 89.38%, 87.17%, and 86.50% respectively for the 70-15-15 split, and 86.03%, 84.37%, and 83.61% respectively for the 60-15-25 split.

In conclusion, the whole image classification performance, as depicted by precision, recall, F1-score, and accuracy, varies based on the number of classes and the specific class combinations. N-related combinations in the 2-class category consistently showed perfect precision, recall, and F1-scores, indicating optimal classification performance. These results provide significant insights and avenues for further research and optimization in whole image classification.

| No. of Classes | Classes | A | P | R | F | | No. of Classes | Classes | A | P | R | F |
|---|---|---|---|---|---|---|---|---|---|---|---|---|
| **4 Class** | **D, P, S, V** | **83.13** | 83.22 | 83.13 | 82.26 | | **4 Class** | **D, P, S, V** | **78.10** | 78.60 | 78.10 | 76.75 |
| **3 Class** | **D, S, V** | **92.64** | 93.48 | 92.64 | 92.54 | | **3 Class** | **D, S, V** | **87.50** | 88.22 | 87.50 | 87.48 |
| | **P, S, V** | 89.83 | 89.71 | 89.83 | 89.73 | | | **P, S, V** | 85.71 | 85.80 | 85.71 | 85.71 |

| | | | | | | | | | | | |
|---|---|---|---|---|---|---|---|---|---|---|---|
| | **D, P, S** | 81.35 | 82.66 | 81.35 | 80.72 | | **D, P, S** | 76.28 | 76.65 | 76.28 | 74.87 |
| | **D, P, V** | 87.30 | 87.30 | 87.30 | 87.30 | | **D, P, V** | 82.69 | 82.52 | 82.69 | 82.30 |
| **2 Class** | **N, D** | **100.0** | 100.00 | 100.00 | 100.00 | **2 Class** | **N, D** | **100.0** | 100.00 | 100.00 | 100.00 |
| | **N, P** | **100.0** | 100.00 | 100.00 | 100.00 | | **N, P** | **100.0** | 100.00 | 100.00 | 100.00 |
| | **N, S** | **100.0** | 100.00 | 100.00 | 100.00 | | **N, S** | **100.0** | 100.00 | 100.00 | 100.00 |
| | **N, V** | **100.0** | 100.00 | 100.00 | 100.00 | | **N, V** | **100.0** | 100.00 | 100.00 | 100.00 |
| | **D, P** | 87.17 | 89.38 | 87.17 | 86.50 | | **D, P** | 84.37 | 86.03 | 84.37 | 83.61 |
| | **D, S** | 95.45 | 95.80 | 95.45 | 95.42 | | **D, S** | 95.83 | 96.13 | 95.83 | 95.81 |
| | **D, V** | **100.0** | 100.00 | 100.00 | 100.00 | | **D, V** | 96.20 | 96.23 | 96.20 | 96.20 |
| | **P, S** | 88.57 | 89.26 | 88.57 | 88.62 | | **P, S** | 82.75 | 83.07 | 82.75 | 82.82 |
| | **P, V** | 89.74 | 91.20 | 89.74 | 89.34 | | **P, V** | 89.23 | 89.34 | 89.23 | 89.08 |
| | **S, V** | 95.45 | 95.80 | 95.45 | 95.42 | | **S, V** | 95.89 | 96.23 | 95.89 | 95.89 |

**Table 5**: AZH Whole image-based classification with different data split (Left - 70%,15%,15%, Right - 60%,15%,25%). P = Precision, R = Recall, F = F1-score, A = Accuracy

## C. Cross Validation

Cross-validation is a robust methodology we employed in our experiments to validate the performance of our machine learning model. It involves splitting the data into several subsets or 'folds', in this case, five. We then train the model on all but one of these folds, testing the model's performance on the unused fold as displayed in **Table 7** . This process is repeated for each fold, giving us a better understanding of how our model might perform on unseen data. It's particularly useful in situations where our dataset is limited, as it maximizes the use of data for both training and validation purposes. Due to resource constraints, our experimental scope was confined to select procedures. As such, we were only able to conduct a limited subset of experiments, focusing on those deemed most crucial and promising.

In the first scenario, we explored an approach called "ROI without location" with an 80-20 data split. Here, the average accuracy varied across different groupings of classes. The accuracy for a grouping of six classes fluctuated between 80.01% to 85.34%, giving an average of 82.58%. For five classes, it varied from 80.71% to 87.14%, with an average of 82.28%. In a group of four classes, we observed a higher average accuracy of 95.65%, while three classes gave us an average accuracy of 74.80%.

The second method we looked at was "ROI with location". Here, we noticed a similar pattern to our first method. The six-class grouping showed an average accuracy of 83.77%, with individual tests ranging from 80.10% to 86.91%. The five-class grouping had an average accuracy of 81.85%, ranging between 78.57% and 84.28%. For four classes, the average accuracy was high again at 95.50%, while the three classes gave us an average of 76.60%.

Finally, we examined the "whole image" method with the same 80-20 data split. A four-class grouping resulted in an average accuracy of 78.34%. One group of three classes managed a much higher average accuracy of 89.86%, while the other three-class group had an average accuracy of 78.22%.

Overall, these results show that the different methods and the number of classes used can have varied impacts on performance.

### D. Robustness and Reliability

To assess the robustness and reliability of our model, we performed multiple tests with varying class distributions on two distinct datasets: the newly created AZH Dataset and the Medetec Dataset. We picked the latter due to its unique data collection and distribution features. Through rigorous testing, we gauged our model's adaptability to diverse conditions.
First, we examined our model on the AZH dataset with a class distribution of 60-15-25 for classes D, P, and V. The model showed notable robustness, achieving an accuracy, precision, recall, and F1-score of 82.69%, 82.52%, 82.69%, and 82.30% respectively.
Next, we used the Medetec dataset for a whole image-based classification. The model continued to showcase excellent robustness despite a different data distribution, registering an accuracy, precision, recall, and F1-score of 87.50%, 87.44%, 87.50%, and 87.43% respectively.
We then altered the class distribution to 70-15-15 on the AZH dataset. The model continued to perform robustly, achieving 87.30% accuracy. In a similar variation on the Medetec dataset, the model held its high performance with accuracy, precision, recall, and F1-score of 88.57%, 88.65%, 88.57%, and 88.50%.
The series of tests reaffirm our model's consistency and adaptability, demonstrating its ability to perform at a high level regardless of class distribution changes or dataset characteristics. This confirms its robustness and versatility as a data analysis tool.

| No. of Classes | Classes | A | P | R | F |
|---|---|---|---|---|---|
| 3 Class | D, P, V | **88.57** | 88.65 | 88.57 | 88.50 |

| No. of Classes | Classes | A | P | R | F |
|---|---|---|---|---|---|
| 3 Class | D, P, V | **87.50** | 87.44 | 87.50 | 87.43 |

**Table 6**: Medetec Whole image-based classification with different data split (Left - 70%,15%,15%, Right - 60%,15%,25%). P = Precision, R = Recall, F = F1-score, A = Accuracy

| ROI without Location with 80-20 Data Split (%) | | | | | | | |
|---|---|---|---|---|---|---|---|
| No. of Classes | Classes | Fold1 | Fold2 | Fold3 | Fold4 | Fold5 | AVG |
| 6 Class | BG, N, D, P, S, V | 80.01 | 80.01 | 82.72 | **85.34** | 84.81 | 82.58 |
| 5 Class | BG, N, D, P, S | 81.42 | 80.71 | 80.71 | 81.42 | **87.14** | 82.28 |
| 4 Class | BG, N, D, V | **96.89** | 93.79 | 96.12 | 95.34 | 96.11 | 95.65 |
| 3 Class | D, P, S | **80.00** | 71.00 | 73.00 | 72.00 | 78.00 | 74.80 |
| ROI with Location with 80-20 Data Split (%) | | | | | | | |
| | Classes | Fold1 | Fold2 | Fold3 | Fold4 | Fold5 | AVG |
| 6 Class | BG, N, D, P, S, V | **86.91** | 83.24 | 80.10 | 83.24 | 85.34 | 83.77 |
| 5 Class | BG, N, D, P, S | 80.71 | 83.57 | 78.57 | 82.14 | **84.28** | 81.85 |
| 4 Class | BG, N, D, V | 94.57 | 95.34 | **96.12** | 95.34 | **96.12** | 95.50 |
| 3 Class | D, P, S | 73.00 | 78.00 | 79.00 | 73.00 | **80.00** | 76.60 |
| Whole Image with 80-20 Data Split (%) | | | | | | | |
| | Classes | Fold1 | Fold2 | Fold3 | Fold4 | Fold5 | AVG |
| 4 Class | D, P, S, V | 77.71 | 78.37 | 79.27 | 77.47 | **78.87** | 78.34 |
| 3 Class | D, S, V | 90.10 | 86.81 | **94.50** | 91.20 | 86.68 | 89.86 |
| 3 Class | D, P, S | 79.74 | 75.94 | **81.01** | 79.74 | 74.68 | 78.22 |

**Table 7**: Cross Validation performed on randomly selected classes for each of the above experiment.

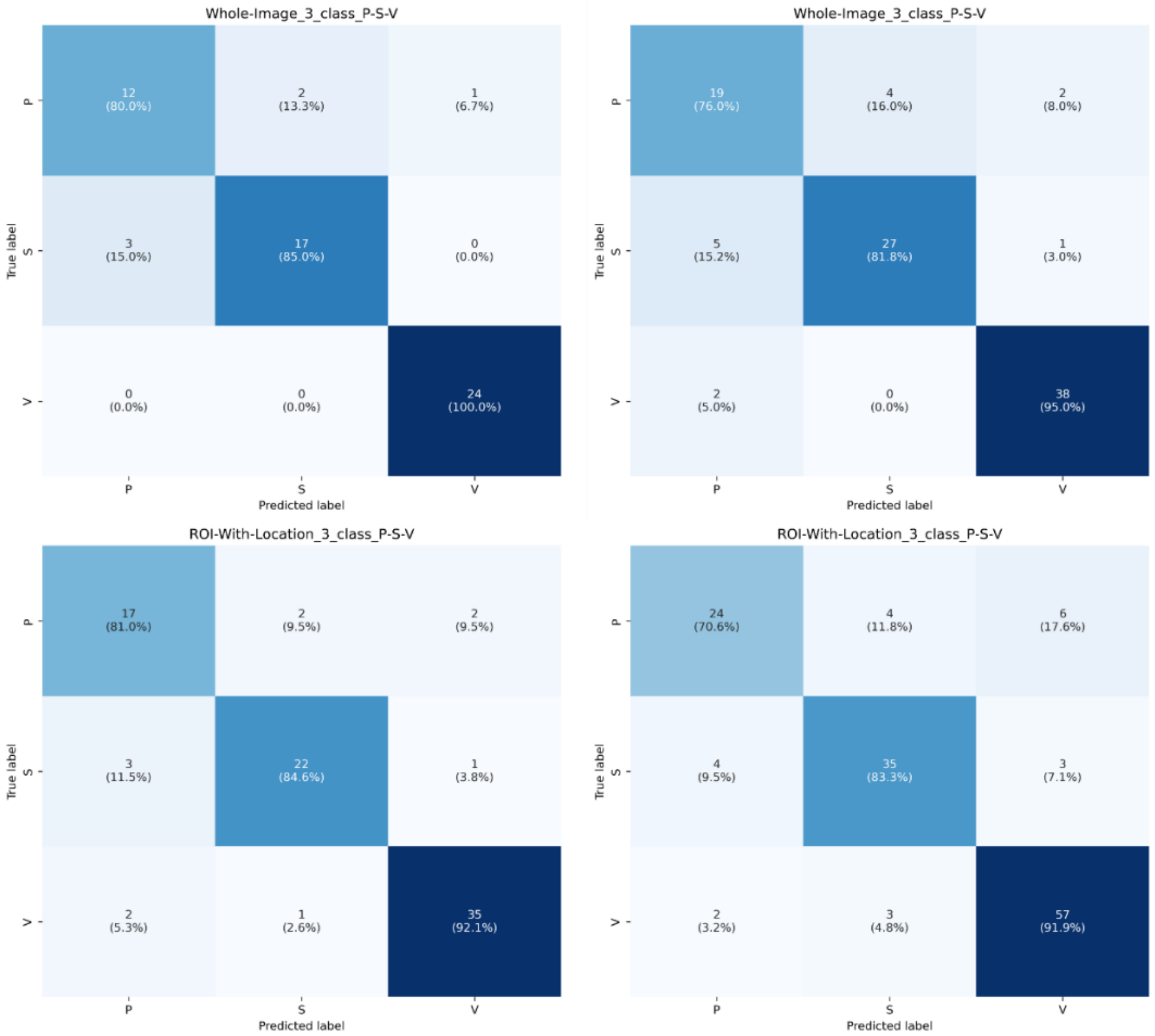

**Figure 8**: Confusion matrix for three class classification on P-S-V, left column displays dataset with 70/15/15 and right column displays dataset with 60/15/25.

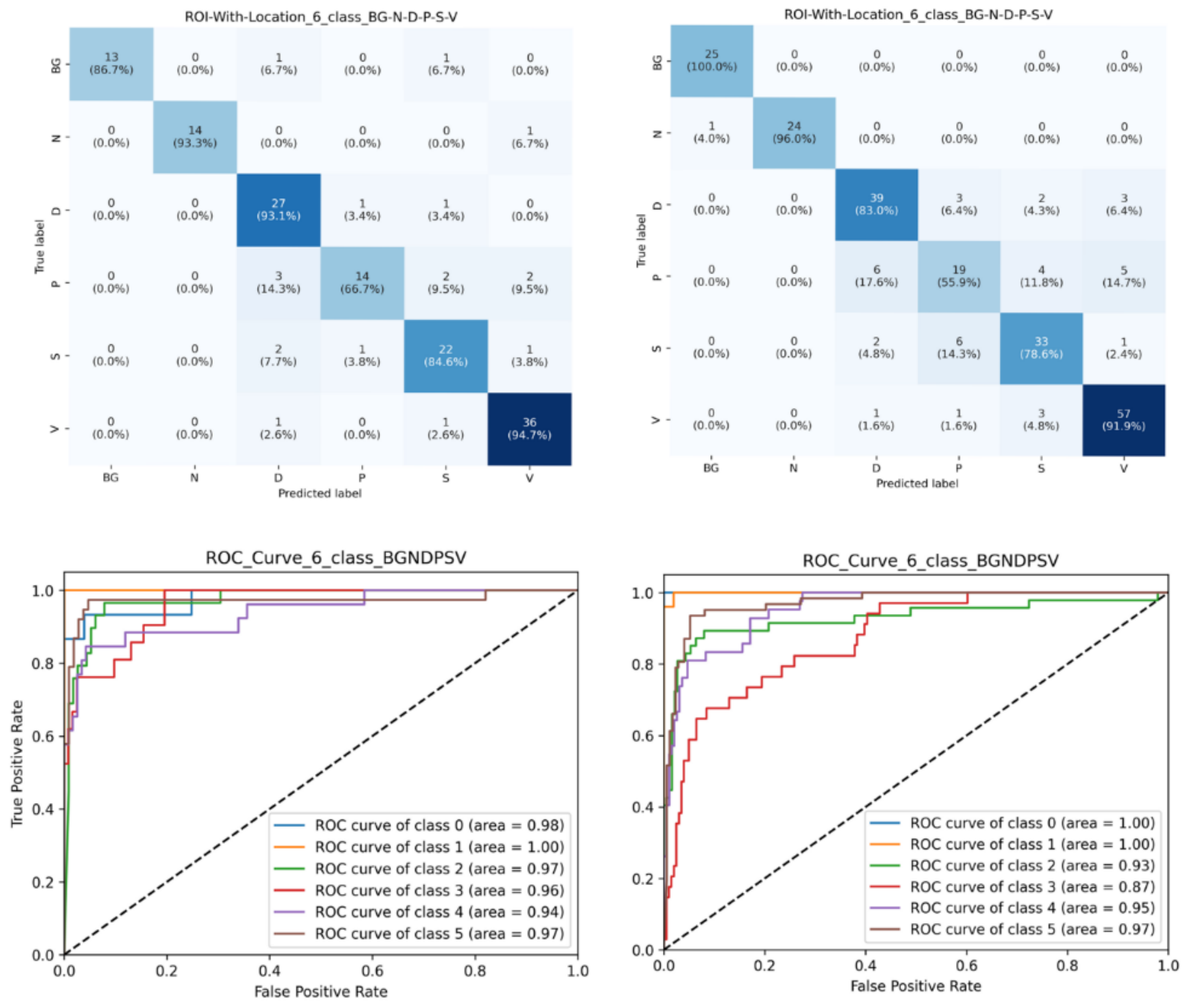

**Figure 9**:Confusion matrix and ROC curve for six class classification on BG-N-D-P-S-V (class 0-1-2-3-4-5), left column displays dataset with 70/15/15 and right column displays dataset with 60/15/25.

| Work | No. of Classes | Classification | Evaluation Metrics | Previous work | | Present work | |
|---|---|---|---|---|---|---|---|
| | | | | Dataset And Split size | Result (%) | Dataset And Split size | Result (%) |
| **Rostami et al. [36]** | **6 Class** | **BG, N, D, P, S, V** | Accuracy | AZH Dataset ROI (NL) (60-15-25) | 68.69 | AZH Dataset ROI (NL) (60-15-25) | **80.42** |
| | **5 Class** | **BG, N, D, P, V** | | | 79.76 | | **83.41** |
| | | **BG, N, D, S, V** | | | 84.94 | | **91.04** |
| | | **BG, N, D, P, S** | | | 81.49 | | **84.39** |
| | | **BG, N, P, S, V** | | | 83.53 | | **87.23** |
| | **4 Class** | **BG, N, D, V** | | | 89.41 | | **96.22** |
| | | **BG, N, P, V** | | | 86.57 | | **91.09** |
| | | **BG, N, S, V** | | | **92.20** | | 91.55 |
| | | **BG, N, D, P** | | | 80.29 | | **87.02** |
| | | **BG, N, D, S** | | | **90.98** | | 89.20 |
| | | **BG, N, P, S** | | | 84.12 | | **85.71** |
| **Anisuzzaman et al.** | **6 Class** | **BG, N, D, P, S, V** | | AZH Dataset ROI (L) | 82.48 | AZH Dataset ROI (L) | **83.82** |
| **[28]** | **5 Class** | **BG, N, D, P, V** | | (60-15-25) (Selected Accuracy based on author's highlight across different models) | 86.46 | (60-15-25) Our model is fixed, and we did not used any different combinations | **89.11** |
| | | **BG, N, D, S, V** | | | 91.00 | | **91.54** |
| | | **BG, N, D, P, S** | | | 83.14 | | **84.39** |
| | | **BG, N, P, S, V** | | | 86.17 | | **88.82** |
| | **4 Class** | **BG, N, D, V** | | | 95.57 | | **96.22** |
| | | **BG, N, P, V** | | | 92.47 | | **93.15** |
| | | **BG, N, S, V** | | | 94.16 | | **96.10** |
| | | **BG, N, D, P** | | | 89.23 | | **89.31** |
| | | **BG, N, D, S** | | | 91.30 | | **93.52** |
| | | **BG, N, P, S** | | | 85.71 | | **88.88** |
| | **3 Class** | **D, S, V** | | | **92.00** | | 90.72 |
| | | **P, S, V** | | | **85.51** | | 84.05 |
| | | **D, P, S** | | | 72.95 | | **73.98** |
| | | **D, P, V** | | | 84.51 | | **86.71** |
| | **2 Class** | **N, D** | | | 100.0 | | **100.0** |
| | | **N, P** | | | **98.31** | | 96.61 |
| | | **N, S** | | | **98.51** | | 98.50 |
| | | **N, V** | | | **100.0** | | 98.85 |
| | | **D, P** | | | 85.00 | | **86.41** |
| | | **D, S** | | | 89.77 | | **89.88** |
| | | **D, V** | | | 94.44 | | **97.24** |
| | | **P, S** | | | **89.47** | | 84.21 |
| | | **P, V** | | | 90.63 | | **92.70** |
| | | **S, V** | | | 97.12 | | **94.23** |
| **Goyal et al. [32]** | **2 Class** | **N, D** | | DFU Dataset | 92.50 | AZH Dataset ROI (L) | **100.0** |
| | | | | | | AZH Dataset ROI (NL) | **100.0** |
| | | | | | | AZH Dataset Whole Image | **100.0** |
| **Aguirre et al. [33]** | **2 Class** | **N, V**<br>**D, V**<br>**P, V**<br>**S, V** | | A dataset of 300 wound images | 85.00 | AZH Dataset ROI (L) | **98.85**<br>**97.24**<br>**92.70**<br>**94.23** |
| | | | | | | AZH Dataset ROI (NL) | **100.0**<br>**97.24**<br>**90.62**<br>**93.26** |
| | | | | | | AZH Dataset Whole Image | **100.0**<br>**96.20**<br>**89.23**<br>**95.89** |

**Table 8**: Previous work comparison. (L – With Location, NL – Without Location)

## Discussion

**A. Comparison with previous work:** Our study presents a comprehensive comparison of our model's performance with those of previous studies, namely the research conducted by Rostami et al. [36], Anisuzzaman et al. [28], Goyal et al. [32], and Aguirre et al.[33]. The comparison is based on accuracy as the evaluation metric, which is a common criterion for classification tasks. For each work, we have tested our model on the same dataset and compared the results as displayed in **Table**

**8**. **Figure 8** and **Figure 9** display confusion matrix for 3-class (P, S and V) and 6-class (BG, N, D, P, S, V). **Figure 9** also display ROC plots for 6-class ROI image with location-based image classification.

In the case of Rostami et al.'s work [36], the classification was carried out on a 6-class, 5-class, and 4-class basis using the AZH dataset. Our model outperformed the previous work by a notable margin across all class divisions. For example, in the 6-class division (BG, N, D, P, S, V), our model improved the accuracy by approximately 11.73%. Furthermore, for the 5-class and 4-class divisions, our model consistently showed improvements, highlighting its efficiency and robustness.

Anisuzzaman et al.'s work [28] also used the AZH dataset, with a focus on 6-class, 5-class, and 4-class divisions. Our model yielded better accuracy results, such as an increase of about 1.34% in the 6-class division. The consistency of improved performance in all divisions showcases the broad applicability of our model.

As for the work of Goyal et al. [32], they only classified into a 2-class division (N, D) using the DFU dataset. When tested on the AZH dataset, our model demonstrated 100% accuracy, similar to their findings. This highlights the versatility of our model in achieving high accuracy across different datasets.

Aguirre et al. [33] conducted their research on a dataset of 300 wound images with a 2-class division. Their model yielded an 85% accuracy rate, while our model, when tested on the AZH dataset, showed a significant improvement in the accuracy rates, ranging from 92.70% to 100%.

**B. Limitations and Future Research:** While our research demonstrates the strengths of our model, it is not without limitations. For instance, all comparisons in our current study were conducted using the AZH and Medetec dataset. Although our model performed commendably on this dataset, the results might not be fully generalizable to all datasets. Hence, the applicability of our model to other datasets remains an area for further investigation.

It's noteworthy that our study does not solely rely on accuracy as the evaluation metric. In an attempt to provide a comprehensive evaluation, we also considered other metrics such as precision, recall, and the F1 score. This thorough approach helps to give a well-rounded understanding of our model's performance. However, despite its strong performance, there could be scenarios or datasets where the model might not yield the same level of success, a potential caveat to be explored in future work.

Future research should be focused on testing the model with larger and more diverse datasets to ensure its generalizability. Specifically, addressing the issue of overlap between healthy and diseased skin, possibly through refining the image preprocessing or feature extraction stages, could yield significant improvements. Furthermore, conducting comparative studies using a wider range of evaluation metrics could offer a broader understanding of the model's strengths and weaknesses. In addition to further empirical evaluation, there is also potential to investigate the theoretical properties of the model. Understanding why the model performs as it does could lead to insights that drive further improvements.

## Conclusion

In this study, we presented a multi-modal wound classification network that uniquely incorporates both images and corresponding wound locations to categorize wounds. Differing from previous research, our approach utilizes a pre-existing body map and two datasets to classify wounds based on their locations. Our model is built on a novel deep learning architecture, featuring parallel squeeze-and-excitation blocks (P_scSE), adaptive gated multi-layer perceptron (MLP), axial attention mechanism, and convolutional layers. The integration of image and location data contributed to superior classification outcomes, demonstrating the potential of multi-modal data utilization in wound management. Despite the benefits, our work has some limitations, including data scarcity which affects the generality of our model.

Looking ahead, future research will aim to enhance our model by incorporating more modalities such as pain level, palpation findings, general observations, wound area and volume, and patient demographics. Addressing data overlaps in wound location will also be a priority to enhance classification accuracy. Our efficient wound care algorithm has significant potential for automation in wound healing systems, offering cost-effectiveness and aiding clinicians in prompt diagnosis and development of suitable treatment plans. Especially in resource-scarce areas, AI-enabled wound analysis can contribute to rapid diagnosis and quality treatment. However, this necessitates proper technical training for both patients and physicians, which will also be a focus of future work. Expanding our dataset will help improve our model's performance and better serve wound care providers and patients alike.

## Data availability

The AZH dataset can be accessed via the following link: (Link). Due to Authorship conflict, we cannot make Medetec dataset public.